\documentclass{article}

\usepackage[preprint]{cs224r_2025}
\usepackage[utf8]{inputenc} 
\usepackage[T1]{fontenc}    
\usepackage{hyperref}       
\usepackage{url}            
\usepackage{booktabs}       
\usepackage{amsfonts}       
\usepackage{nicefrac}       
\usepackage{microtype}      
\usepackage{xcolor}         
\usepackage{amsmath}
\usepackage{graphicx}       
\usepackage{lipsum}         
\usepackage{graphicx}
\usepackage{subfig}
\usepackage[justification=centering]{caption}

\title{Step-wise Policy for Rare-tool Knowledge (SPaRK): Offline RL that Drives Diverse Tool Use in LLMs}

\author{%
  Gabriel Bo \\
  Department of Computer Science\\
  Stanford University\\
  \texttt{gabebo@cs.stanford.edu} \\
  \And
  Koa Chang \\
  Department of Computer Science\\
  Stanford University\\
  \texttt{koachang@stanford.edu} \\
  \And
  Justin Gu \\
  Department of Computer Science\\
  Stanford University\\
  \texttt{justingu@stanford.edu} \\
}

\begin{document}

\begin{titlepage}
\begin{center}
\Large\textbf{Extended Abstract}
\end{center}
\vspace{10pt}

\paragraph{Motivation} Large language models (LLMs) have shown remarkable reasoning capabilities, yet current approaches rely heavily on brute-force sampling or temperature-based exploration, failing to systematically leverage diverse tools available to modern language agents. While recent models like o3 and DeepSeek R1 demonstrate the power of inference-time computation, they tend to converge on familiar patterns rather than exploring the full spectrum of available resources. We address this gap by introducing Step-wise Policy for Rare-tool Knowledge (SPaRK), a framework that explicitly models tool selection as a learnable policy within a multi-step Markov Decision Process, incorporating a composite reward function that incentivizes both correctness and tool diversity.

\paragraph{Method} SPaRK builds on recent advances in step-wise reinforcement learning by introducing a dual-objective reward system that simultaneously optimizes for answer quality and tool diversity. We train a Llama-3.1 8B model through offline Proximal Policy Optimization (PPO) on synthetically generated trajectories from the MMLU-Pro dataset. Our approach uniquely employs a rarity-first exploitation strategy where a GPT-4o judge scores candidate actions across eight distinct tools plus chain-of-thought reasoning. The policy favors less-frequently used but still viable tools (scoring above a threshold of 6.0) to encourage systematic exploration, selecting the lowest-scoring tool among those that pass the quality threshold rather than always choosing the highest-scoring option.

\paragraph{Implementation} We generate synthetic tool-rich trajectories from 2,500 MMLU-Pro questions, rolling each for 5 steps. At each step, a Llama-3 8B-Instruct generator produces intermediate thoughts and a GPT-4o judge scores 9 candidate actions (8 tools + CoT) on a 0-10 scale. The PPO implementation uses QLoRA-adapted models with a clipped surrogate objective augmented by KL divergence regularization to maintain training stability. We train separate actor and critic networks with LoRA rank 8, using a learning rate of 1e-5 and clip parameter of 0.2, processing batches of 8 subtrajectories over 4 epochs.

\paragraph{Results} Our evaluation on MMLU-Pro demonstrates substantial performance improvements: the baseline Llama-3.1 8B achieves 22.4$\%$ accuracy, supervised fine-tuning improves to 26.2$\%$, PPO without tool diversity reaches 33.0$\%$, while SPaRK achieves 40.8$\%$ accuracy. This 7.8 percentage point improvement over standard PPO validates that explicitly incorporating tool diversity into the training objective yields significant gains. The generated dataset of 12,500 fully annotated steps captures rich intermediate supervision, tracking both correctness and exploration quality across diverse reasoning trajectories.

\paragraph{Discussion} SPaRK's key innovation lies in unifying exploration and exploitation through a rarity-first strategy that encourages choosing the least-used tool among viable options, preventing convergence on high-frequency tools like chain-of-thought. Unlike existing approaches that separate these objectives, our minimum-score-above-threshold rule enables systematic exploration without sacrificing utility. Our results suggest that learned exploration policies can achieve competitive accuracy while maintaining significantly higher entropy in tool selection.

\paragraph{Conclusion} SPaRK demonstrates that algorithmic exploration through explicit tool diversity can enhance reasoning capabilities without sacrificing accuracy, nearly doubling the performance of supervised fine-tuning on MMLU-Pro. By teaching models not just what to think but how to systematically explore diverse reasoning strategies, our work challenges assumptions about model scale and temperature-based sampling as primary drivers of improvement. Instead, highlighting principled exploration policies over tools may be a promising direction for advancing language model capabilities in complex, multi-domain reasoning tasks.

\end{titlepage}

\maketitle

\begin{abstract}
We present Step-wise Policy for Rare-tool Knowledge (SPaRK), a novel reinforcement learning framework that teaches large language models to explore diverse tool usage patterns beyond conventional high-temperature sampling. Building on recent advances in step-wise reinforcement learning, we introduce a dual-objective reward system that simultaneously optimizes for answer quality and tool diversity, training a Llama-3.1 8B model through offline PPO on synthetically generated trajectories from the MMLU-Pro dataset. Our approach uniquely employs a rarity-first exploitation strategy where a GPT-4o judge scores candidate actions across eight distinct tools plus chain-of-thought reasoning, with the policy favoring less-frequently used but still viable tools to encourage systematic exploration. Empirical results demonstrate that SPaRK achieves competitive performance across 14 MMLU-Pro categories while exhibiting significantly higher entropy in tool selection compared to both baseline and supervised fine-tuning approaches, suggesting that algorithmic exploration through explicit tool diversity can enhance reasoning capabilities without sacrificing accuracy.
\end{abstract}

\section{Introduction}
The rapid advancement of large language models (LLMs) has sparked considerable interest in enhancing their reasoning capabilities through tool augmentation and multi-step decision making. Recent breakthroughs from models like Gemini 2.5 Pro, OpenAI's o3 and o4, and DeepSeek R1 \citep{gemini2025blog}\citep{openai2024gpt4technicalreport}\citep{deepseek_r1} have demonstrated the power of inference-time computation and reinforcement learning for complex reasoning tasks \citep{deepseek_r1}. However, these approaches often rely on brute-force sampling or temperature-based exploration, which fails to systematically leverage the diverse tools available to modern language agents.

Our work addresses this gap by introducing Step-wise Policy for Rare-tool Knowledge (SPaRK), a framework that explicitly models tool selection as a learnable policy within a multi-step Markov Decision Process. Unlike existing methods that optimize solely for answer quality, SPaRK incorporates a composite reward function that incentivizes both correctness and tool diversity. By treating each reasoning step as a discrete action choice among multiple tools—including calculators, dataset lookups, web search, and various computational interfaces—we enable the model to learn a distribution over tool usage that goes beyond random exploration. This approach is particularly motivated by the observation that many reasoning tasks benefit from heterogeneous information sources, yet current models tend to converge on familiar patterns rather than exploring the full spectrum of available resources.

We validate SPaRK on the challenging MMLU-Pro dataset \citep{wang2024mmluprorobustchallengingmultitask}, which spans 14 subjects requiring specialized expertise and multi-step reasoning. Our experiments compare four configurations: a baseline Llama-3.1 8B-Instruct model \citep{llama31_8b}, a supervised fine-tuned variant trained on question-answer pairs, a PPO-based model that emulates SWiRL, and our SPaRK model trained via offline PPO on synthetic tool-augmented trajectories \citep{schulman2017proximalpolicyoptimizationalgorithms}. The results demonstrate that learned exploration policies can achieve competitive accuracy while maintaining significantly higher entropy in tool selection, suggesting that systematic diversity in tool usage represents an underexplored dimension for improving language model reasoning. This work contributes evidence that decoupling exploration from exploitation in tool selection can yield statistically significant improvements in multi-domain reasoning tasks.

\section{Related Work}
\subsection{Multi-step Reinforcement Learning}
Recent advances in multi-step reinforcement learning for language models have followed several promising directions. The SWiRL framework by \citet{goldie2025syntheticdatageneration} pioneered synthetic step-wise reinforcement learning by decomposing tool-use traces and applying per-step rewards, achieving double-digit gains on GSM8K and HotpotQA \citep{gsm8k2021}\citep{yang2018hotpotqa}. Similarly, inference-time sampling approaches like Large Language Monkeys demonstrate that repeated sampling with high temperature can improve pass-at-k accuracy log-linearly. However, both approaches optimize primarily for answer quality, with exploration occurring implicitly through temperature-based sampling rather than learning explicit distributions over tool choices. Our work extends SWiRL's step decomposition methodology but introduces a critical innovation: a reward function that explicitly values tool diversity alongside correctness, enabling the model to discover novel solution pathways through systematic exploration of underutilized tools.

\subsection{Exploration Post-Training}
The challenge of balancing exploration and exploitation in post-training has been addressed from different angles. \citet{liu2020decouplingexplorationmeta} proposed DREAM, which decouples task-relevant exploitation from exploration through mutual-information rewards, demonstrating that separating these objectives can improve meta-reinforcement learning performance. More recently, \citet{feng2025retoolreinforcementlearningstrategic} introduced ReTool, which uses reinforcement learning for strategic tool selection but focuses on efficiency rather than diversity. While these works advance our understanding of exploration in language models, none simultaneously learn exploration policies that select among heterogeneous tools while rewarding both answer quality and trajectory diversity. SPaRK bridges this gap by joining SWiRL-style step decomposition with DREAM's decoupled objectives, creating a value-aware tool-selection policy that treats rare tool usage as a critical and utilizable feature, ultimately demonstrating that controlled exploration through tool diversity can enhance reasoning without sacrificing accuracy.

\section{Method}

\subsection{Baseline: Supervised Fine-Tuning (SFT)}
\begin{figure}[t]
  \centering
  \includegraphics[width=0.85\linewidth]{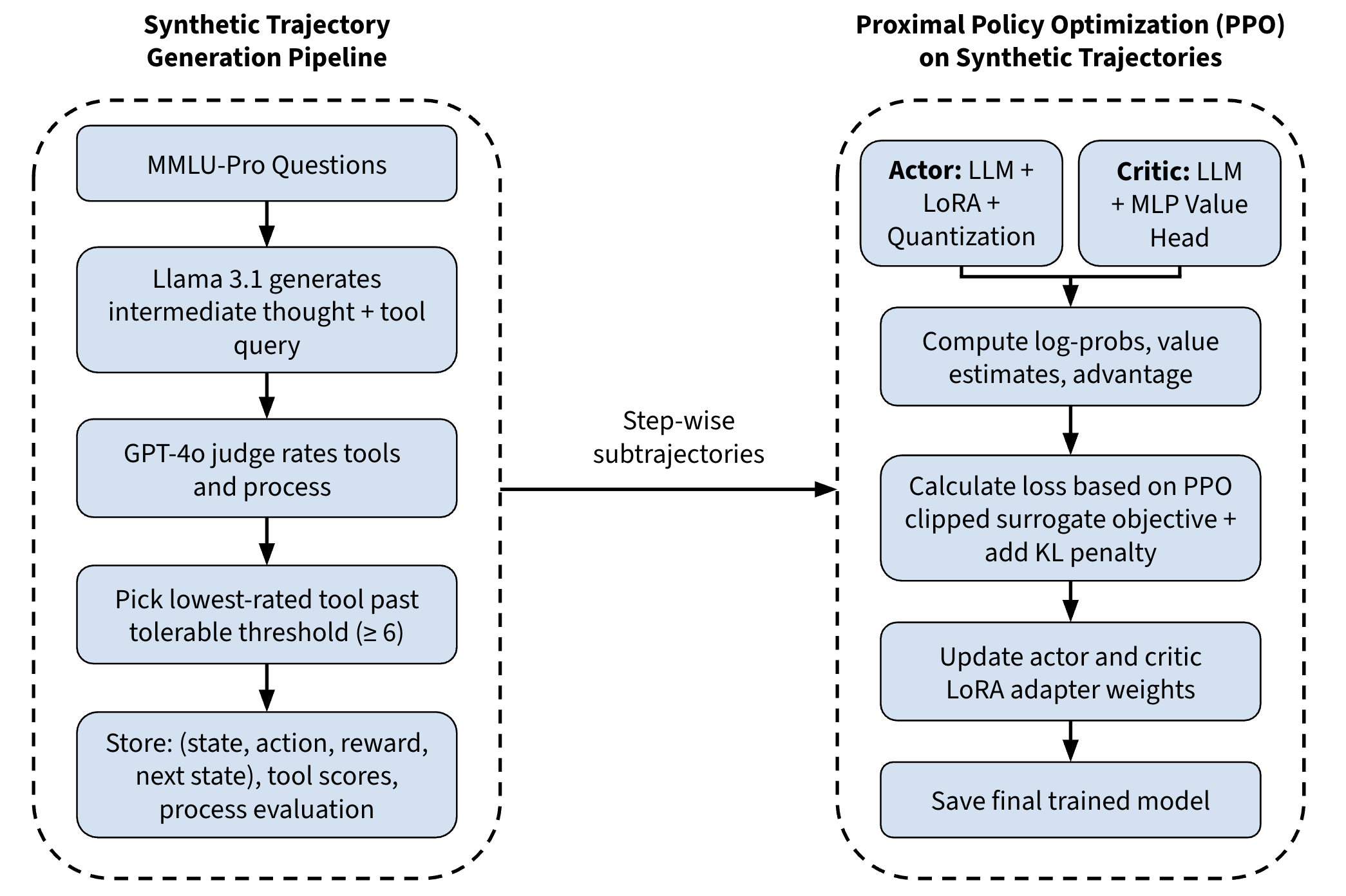}
  \caption{SPaRK Method Overview, from synthetic data generation to PPO-based training.}
  \label{fig:results}
\end{figure}

As a reference baseline we perform straightforward supervised fine-tuning (SFT) of Meta-Llama-3.1 8B-Instruct-Reference \citep{grattafiori2024llama3herdmodels} on the original MMLU-Pro training split. Each example is converted to JSONL with a short system message, a user turn containing the question plus its four options, and an assistant turn that holds only the correct letter. We train with lightweight LoRA adapters on the frozen base model (no RL, no tools), then run inference on the MMLU-Pro test set by prompting the model to “think step-by-step and output the final choice in <answer>{A–J}</answer>.” A small parser model (Llama-3 8B-Turbo) extracts the letter from the tag, and accuracy is computed directly against the ground-truth key; this yields a clean single-number benchmark from the results in Figure \ref{fig:results} against which the tool-augmented PPO variant can be compared. The model accuracies are shown for both the baseline without finetuning and with a simple SFT for the 14 categories in MMLU-Pro. SFT was trained on LoRA with a resulting rank of 64 with a learning rate of $1e{-5}$.

\subsection{Synthetic Data Generation}
We first generate a synthetic, tool-rich dataset off of the data from Tiger Lab’s MMLU-Pro \citep{wang2024mmluprorobustchallengingmultitask}. This dataset consists of trajectories on which reinforcement learning can be used to train a Llama-3.1 8B-Instruct pre-trained model \citep{grattafiori2024llama3herdmodels}. To make the trajectories, a stronger GPT-4o \citep{openai2024gpt4technicalreport} judge scores nine candidate actions (eight tools + chain-of-thought (CoT)), and the rewards favor answers that call less-used tools. Each step becomes a labeled (state, action, reward) tuple for offline PPO \citep{schulman2017proximalpolicyoptimizationalgorithms}, trained via LoRA \citep{hu2021loralowrankadaptationlarge}and shallow unfreezing. 

\begin{figure}[h]
    \centering
    \includegraphics[width=1\linewidth]{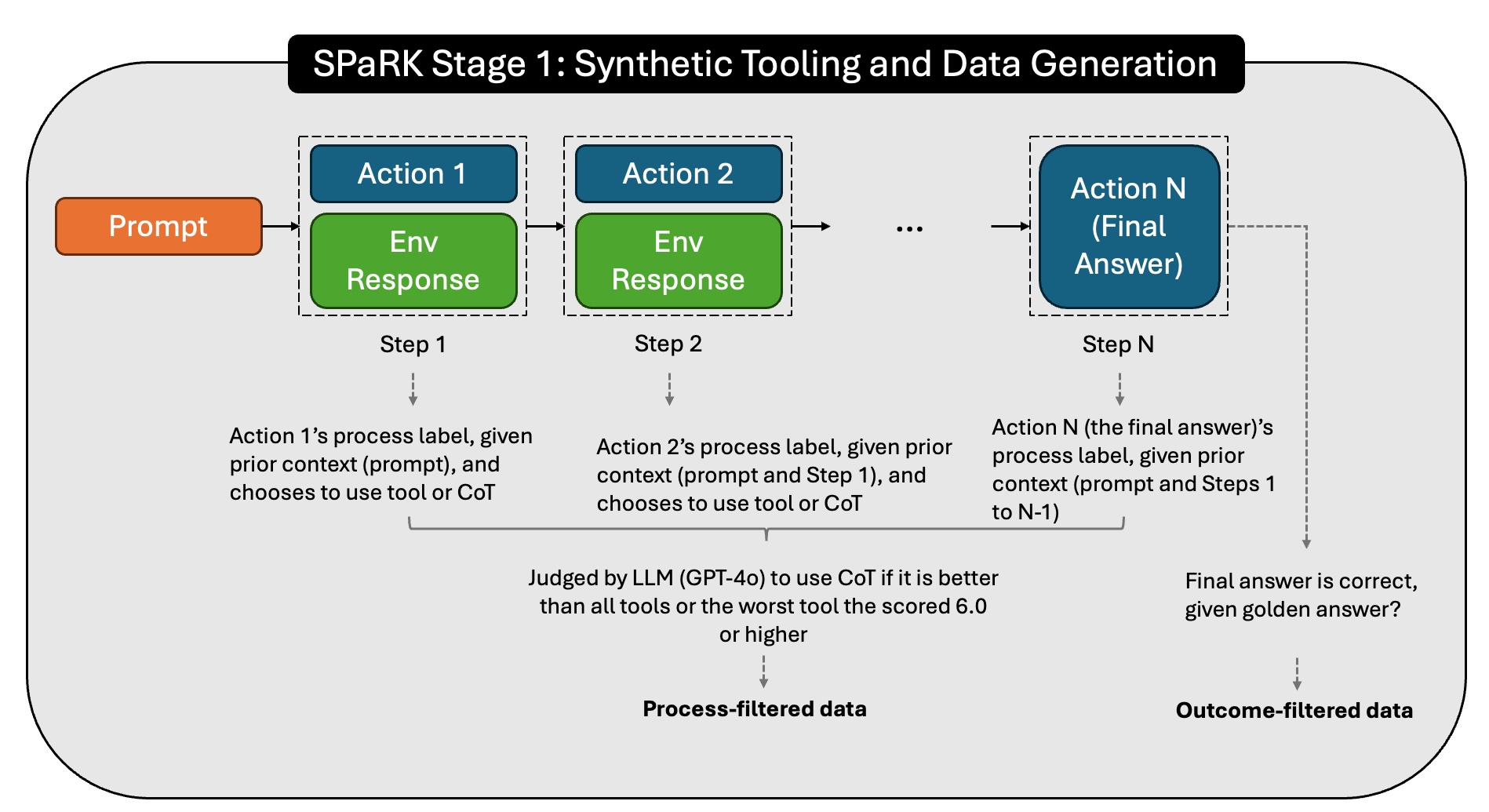}
    \caption{SPaRK Synthetic Data Generation Process}
    \label{fig:spark1}
\end{figure}

\subsection{Reinforcement Learning: Proximal Policy Optimization (PPO)}

We implement Proximal Policy Optimization (PPO) with QLoRA-adapted large language models (LLMs) \citep{large_language_monkeys} for reinforcement learning over natural language reasoning tasks. Our models are built using Hugging Face's Transformers library, enabling modular integration of quantized LLMs with lightweight LoRA (Low-Rank Adaptation) adapters for efficient fine-tuning.

In line with Actor-Critic algorithms, we define two classes an Actor model and a Critic model. The Actor loads a base Hugging Face language model with 4-bit quantization and applies LoRA adapters to fine-tune only a small set of parameters. It is trained to maximize the expected return using PPO and outputs log-probabilities for sampled actions. The Critic consists of a similar base architecture but adds a value head, implemented as a multilayer perceptron (MLP), which estimates the expected return of a given state.

Our PPO implementation follows the clipped surrogate objective introduced by \citet{schulman2017proximalpolicyoptimizationalgorithms}:

\[
L^{\text{CLIP}}(\theta) = \mathbb{E}_t \left[ \min\left( r_t(\theta) \hat{A}_t, \ \text{clip}(r_t(\theta), 1 - \epsilon, 1 + \epsilon) \hat{A}_t \right) \right],
\]

where $r_t(\theta) = \frac{\pi_\theta(a_t \mid s_t)}{\pi_{\theta_{\text{old}}}(a_t \mid s_t)}$ denotes the ratio of probabilities under the new and old policies, and $\hat{A}_t$ is the estimated advantage at timestep $t$. The clipping operation constrains updates that would otherwise cause large policy shifts, effectively penalizing updates that over-exploit positive advantages while allowing beneficial updates when the advantage is negative. This results in a more stable and conservative policy update.

\begin{figure}
    \centering
    \includegraphics[width=1\linewidth]{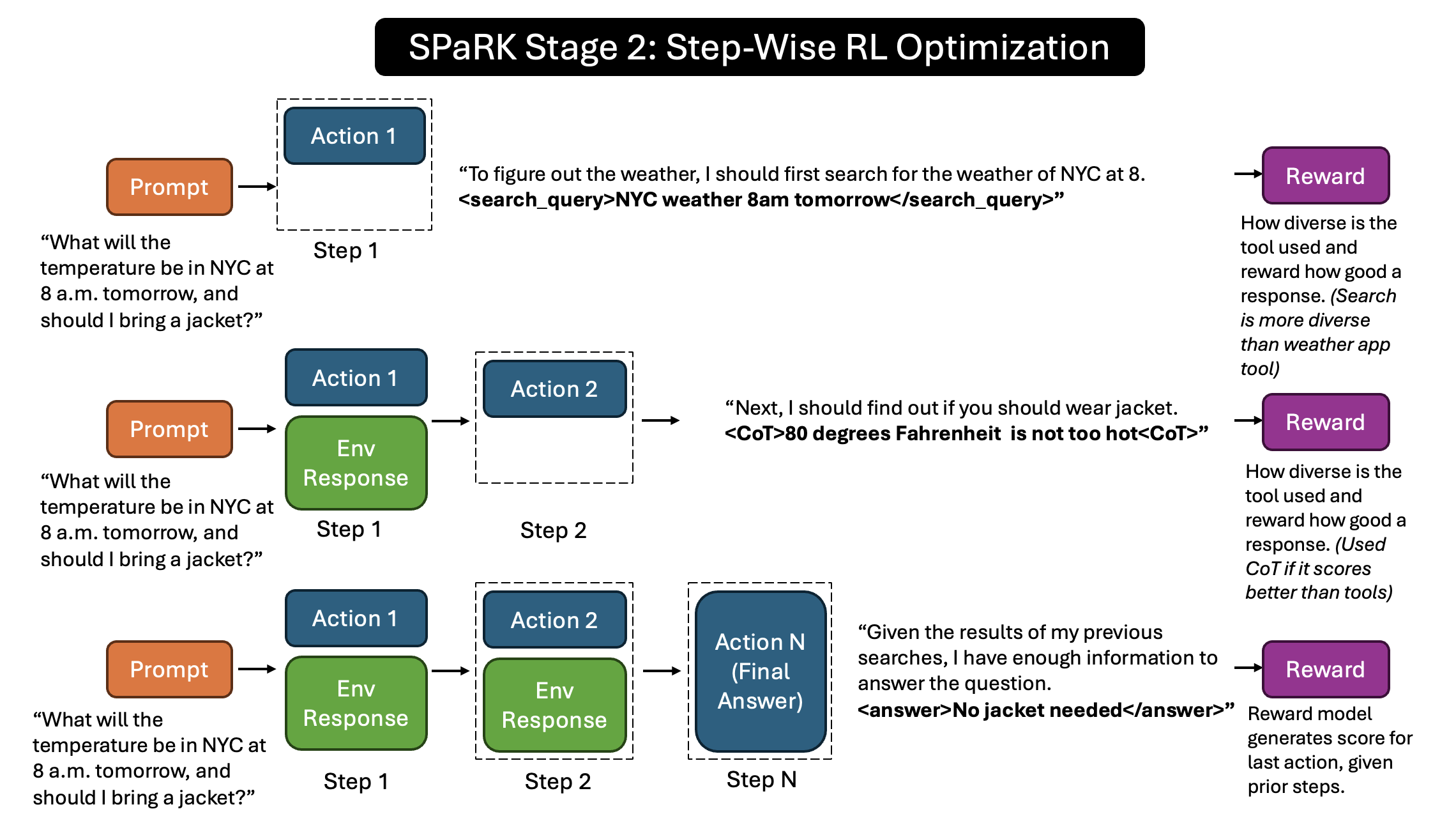}
    \caption{SPaRK Reinforcement Learning Optimization Procedure}
    \label{fig:enter-label}
\end{figure}

We utilize a simple reward model to incentivize exploration. For each subtrajectory, we calculate the reward as the difference between the scores of the chosen action and the best action, as evaluated by the judge LLM in the data generation. Thus, subtrajectories with larger rewards indicate greater exploration. To judge this exploration, while not being subject to the model performing too poorly, we also added a factor of whether the "process was ok" judged by the same judge-LLM. It was then varied based on $\rho$, creating a convex equation that one could choose if they favored exploration or exploitation more. This is what makes it similar to the paper published by \citet{goldie2025syntheticdatageneration}, but takes into higher consideration the tooling that's chosen.

\[R_t = \rho(\text{Best Score} - \text{Chosen Score}) - (1-\rho)(\text{process\_ok})\]

Due to our PPO implementation being offline using the synthetically generated dataset, our advantage estimate is derived by subtracting the values, as calculated by the Critic model with pre-updated parameters, from the rewards:

\[\hat{A_t} = R_t - \hat{V}_{\phi_{old}}(s_t)\]

To maintain training stability, we incorporate KL divergence regularization into our PPO objective function following established practices in policy gradient methods \citep{schulman2017proximalpolicyoptimizationalgorithms}. We augment the standard PPO clipped surrogate objective with an additional KL penalty term:

\begin{equation}
L_{\text{total}} = L^{\text{CLIP}} + \beta \cdot L_{\text{KL}}
\end{equation}

where $\beta$ is the KL coefficient (default: 0.1) and $\mathcal{L}_{\text{KL}}$ represents the KL divergence penalty. To approximate the KL divergence, we use the squared difference of log probabilities:

\begin{equation}
L_{\text{KL}} = \mathbb{E}\left[(\log \pi_\theta(a|s) - \log \pi_{\theta_{\text{old}}}(a|s))^2\right]
\end{equation}

This quadratic approximation is valid for small policy updates and provides stronger regularization compared to the linear KL divergence, helping to prevent large policy shifts that could destabilize training.

We also implement early stopping when the approximate KL divergence exceeds the target KL threshold (default target: 0.2). This prevents excessive policy deviation within a single training iteration, which aids further in balancing exploration with stability. This allows the policy to adapt while preventing unrestricted policy gradient updates in offline reinforcement learning settings.

We chose PPO as our initial RL algorithm because it balances stability, simplicity, and compatibility with offline learning. The clipped surrogate objective is resistant to large policy updates, which helps when training on a fixed dataset of pre-collected tool-use trajectories. Unlike other algorithms that may require online rollouts or are sensitive to reward shaping, PPO integrates naturally with our composite reward function that promotes both answer correctness and exploration of rare tools. It is also relatively straightforward, which allowed us to quickly implement and test our methodology, making it ideal for evaluating the effectiveness of our exploration-focused approach before extending to more complex RL methods.

This PPO objective composed of the clipped surrogate objective with KL penalty is utilized to update the actor (policy network). We update the critic separately using a standard mean-squared error loss between the estimated value using the current critic parameters and the empirical reward:

\[
L_{critic}(\theta) = \mathbb{E}_t \left[ (\hat{V}_{\phi} (s_t) - R_t)^2 \right]
\]

\section{Experimental Setup}
\subsection{Synthetic Data Generation}
We harvest 2,500 questions from MMLU-Pro \citep{wang2024mmluprorobustchallengingmultitask} and roll each for $K = 5$ steps.  At every step, a Llama-3 8B-Instruct generator ``thinks aloud,'' embeds an \texttt{\textless intended\_query\_for\_tool\textgreater} tag, and hands the resulting state to a GPT-4o judge that scores a fixed set of 9 candidate actions 8 tools and CoT) on a 0 to 10 scale.  Unlike the \citet{goldie2025syntheticdatageneration} paper that inspired the synthetic generation of this paper, we create a reward pattern not just based off of the judge LLM on the quality of the state, but crafting an action space that enables choosing tools.
The policy follows rarity-first exploitation: among tools with a score greater than a cutoff threshold, it picks the \(\arg\min\) (lowest-scoring) one, forcing exploration of
less-favored resources. We empirically determined the threshold to be $6.0$; starting with $5.0$ created many trajectories with incorrect answers, and increasing the threshold to $6.0$ yielded effective results. However, if the judge LLM deems the CoT action strictly superior compared to every tool, it executes CoT instead.  
The judge-LLM would also rank whether the step, given the context, was a reasonable step and mark a binary $\text{process\_ok}$ as \texttt{True} or else \texttt{False}. This was done to ensure that the tools diversity wasn't the only thing we were basing the RL training off of, but the quality of the samples would be filtered out in the process as well. We would also filter out if the out-come was correct or not, similar to what \ref{fig:spark1} displays.
The executed action’s score becomes the step reward
(\(\textit{reward\_raw} = \textit{chosen\_score}\)); the highest alternative
(\(\textit{best\_score}\)) is logged for diagnostics.  
This procedure yields 12,500 fully traced agent steps and a compact
\texttt{sub\_trajectories} table of \((\text{state},\text{action},\text{reward},\text{next state})\) tuples, described in more detail in Section 5.1.

\subsection{PPO Implementation}

The PPO agent is initialized with the actor and critic models, the tokenizer, and a set of hyperparameters. The actor and critic models were trained on LoRA configurations with a rank of 8, alpha of 16, dropout of 0.05. The PPO trainer utilizes a learning rate of $1e{-5}$ and a clip parameter of 0.2, which aligns with the original value proposed by \cite{schulman2017proximalpolicyoptimizationalgorithms} for the clipped surrogate objective. During training, batches of trajectory data are tokenized and reward signals are computed. For each batch of 8 subtrajectories, we compute the log-probabilities and value estimates for the sampled actions, derive the generalized advantage estimates, and perform gradient updates for both the actor and critic using the PPO objective.

Our training pipeline begins by loading and configuring the model environment, including setting random seeds for reproducibility and preparing datasets. We pass in the dataset of synthetic subtrajectories from the Hugging Face repository, which is validated and converted into a standardized format for efficient processing. After initializing the model and applying quantization for memory efficiency, we train the PPO agent for 4 epochs and log key metrics for later analysis.

\subsection{Evaluation}

Due to the multiple-choice nature of MMLU-Pro, we used accuracy to evaluate our models. For each model, we run it on a test dataset made up of 840 multiple-choice tasks that were set apart from the training dataset. The models generate responses to each input prompt for each task, and we compare these outputs with the correct answers to calculate accuracy.

\section{Results}
\subsection{Synthetic Data Generation}
The generated sub-trajectories power our offline PPO phase. This dataset was saved onto an open-source Huggingface repository, and a sample of the synthetic tool-diverse trajectory looks like this:
\begin{table}[h]
\centering
\small
\setlength{\tabcolsep}{4pt}
\begin{tabular}{l p{6cm} c c c c}
\toprule
\textbf{qid} & \textbf{thought (trimmed)} & \textbf{step} & \textbf{tool} & \textbf{chosen} & \textbf{best} \\
\midrule
531\_089 & ``Need Titan’s mass to compute surface gravity …'' & 1 & search        & 6.2 & 7.5 \\
531\_089 & ``Now plug values into $g=Gm/r^2$ …''             & 2 & python\_repl  & 6.0 & 6.0 \\
531\_089 & ``Final Answer: (B)''                              & 3 & cot           & 8.1 & 8.1 \\
\bottomrule
\end{tabular}
\caption{Example synthetic rollout generated with the MMLU-Pro data (https://huggingface.co/datasets/gabrielbo/swirl-trajectories-mmlu-pro)}
\end{table}

Each trajectory captures a multi-step interaction where the language model formulates intermediate thoughts, selects from a set of available tools, and observes the results, all while being scored by the GPT-4o judge. At every step, the dataset records the model’s internal reasoning, the outcome of its chosen tool action, and the reward derived from the quality of the selected action relative to the best alternative.
Each step also includes the GPT-4o judge's assessment of whether the model’s reasoning and tool usage process were coherent enough to be considered valid for training. This quality check helps ensure the reliability of the offline reinforcement learning signal. Additionally, the final step of each trajectory indicates whether the model ultimately selected the correct multiple-choice answer, allowing for downstream analysis of both reasoning quality and task accuracy.

\begin{figure}
    \centering
    \includegraphics[width=0.85\linewidth]{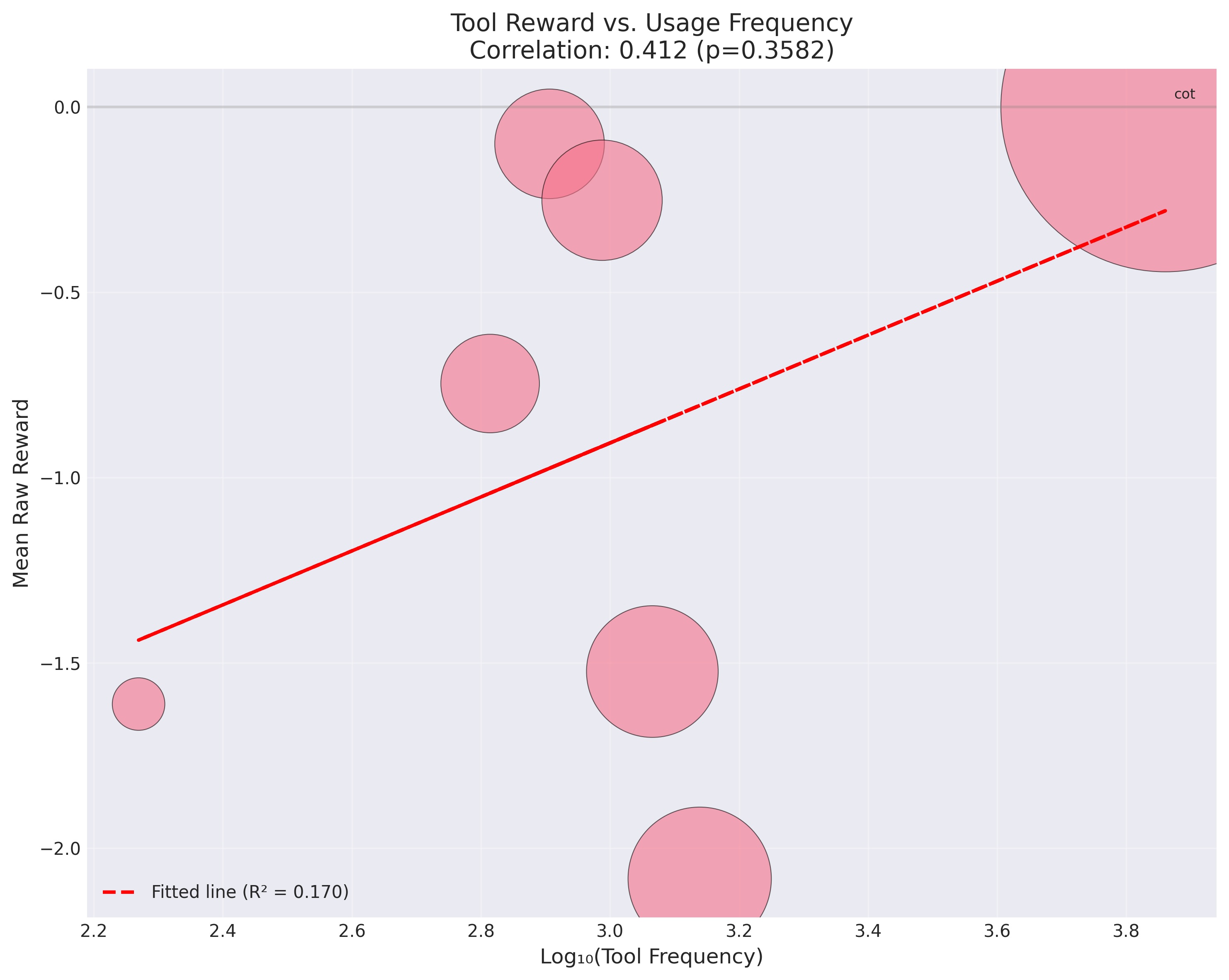}
    \caption{Chain-of-Thought (CoT) frequently had the highest reward and usage. To focus on tool-usage, we examined the distribution of the other tools excluding CoT.}
\end{figure}

The dataset includes 12,500 fully annotated steps across thousands of questions, and is structured to support efficient offline PPO training. By capturing rich intermediate supervision and explicitly tracking both correctness and exploration quality, the dataset enables learning policies that go beyond simple outcome prediction and instead model the process of reasoning itself.

\subsection{Quantitative Evaluation}
We evaluated four model variants on the MMLU-Pro test set: (1) the base Llama-3.1 8B model, (2) a supervised fine-tuned (SFT) model, (3) a PPO-trained model without tool diversity, and (4) our full PPO-trained model with tool-augmented trajectories. Evaluation was based on raw accuracy, the percentage of multiple-choice questions answered correctly across all 14 subject areas in MMLU-Pro.

We observe a clear performance progression across the four evaluated model variants. The base Llama-3.1 8B model achieved 22.4\% accuracy on the MMLU-Pro test set, while supervised fine-tuning (SFT) modestly improved this to 26.2\%. A PPO-trained model that learns from tool-augmented trajectories but deterministically chooses the highest-scoring tool at each step reaches 33.0\% accuracy, verifying that reinforcement learning with tool-use yields improvements in reasoning capabilities. However, our full SPaRK model, which trains on synthetic trajectories to explore rare but viable tools using a reward that balances correctness and diversity, achieves the highest accuracy at 40.8\%. These results support our central claim: reinforcement learning alone improves reasoning over standard finetuning, but explicitly incorporating tool diversity into the training objective leads to even greater gains. The 7.8 percentage point improvement from PPO without diversity to SPaRK underscores that learning a step-wise policy that balances exploration and exploitation can lead to better generalization and problem-solving capability across a broad range of subject areas in MMLU-Pro.

\begin{table}[h]
  \label{tab:performance}
  \centering
  \begin{tabular}{lccc}
    \toprule
    \textbf{Method} & \textbf{Accuracy} \\
    \midrule
    Baseline Llama-3.1 8B & 0.224 \\
    Baseline with SFT & 0.262 \\
    PPO without tool diversity & 0.330 \\
    SPaRK PPO-trained model & 0.408 \\
    \bottomrule
  \end{tabular}
  \caption{Performance Comparison}
\end{table}

\subsection{Qualitative Analysis}
The improvement in performance with SPaRK supports our central hypothesis that encouraging diverse tool usage leads to more effective and generalizable reasoning strategies. While supervised fine-tuning rewards the reproduction of known answers from surface patterns, reinforcement learning optimizes for a sequence of actions that contribute meaningfully to solving a problem. This allows the model to improve reasoning capabilities by engaging more deeply with intermediate steps, learning when and how to apply different tools based on the evolving context of the question.

The MMLU-Pro benchmark spans a wide range of subjects, including science, mathematics, law, and medicine, many of which require multi-step logic, factual retrieval, or symbolic computation. In these domains, naively selecting the most frequent or familiar reasoning path is often insufficient. Without guidance, models tend to rely on dominant patterns observed during pretraining or fine-tuning, which may not translate well to high-complexity, multi-domain reasoning tasks. The PPO training process counters this by teaching the model to distinguish between tools based on their utility for specific subtasks, encouraging greater adaptability across subject areas.

The synthetic trajectories used for offline PPO provide a critical learning signal by including not only the chosen tool and its outcome, but also alternative actions and their relative quality as judged by a stronger model. This comparative supervision helps the policy model learn from suboptimal choices and develop preferences for tools that are useful but underutilized. For example, the model might initially favor chain-of-thought reasoning due to its frequency, but through repeated exposure to trajectories where a search or calculator tool yields better results, it begins to incorporate these tools more appropriately into its decision-making.

\begin{figure}
    \centering
    \subfloat{{\includegraphics[width=0.44\linewidth]{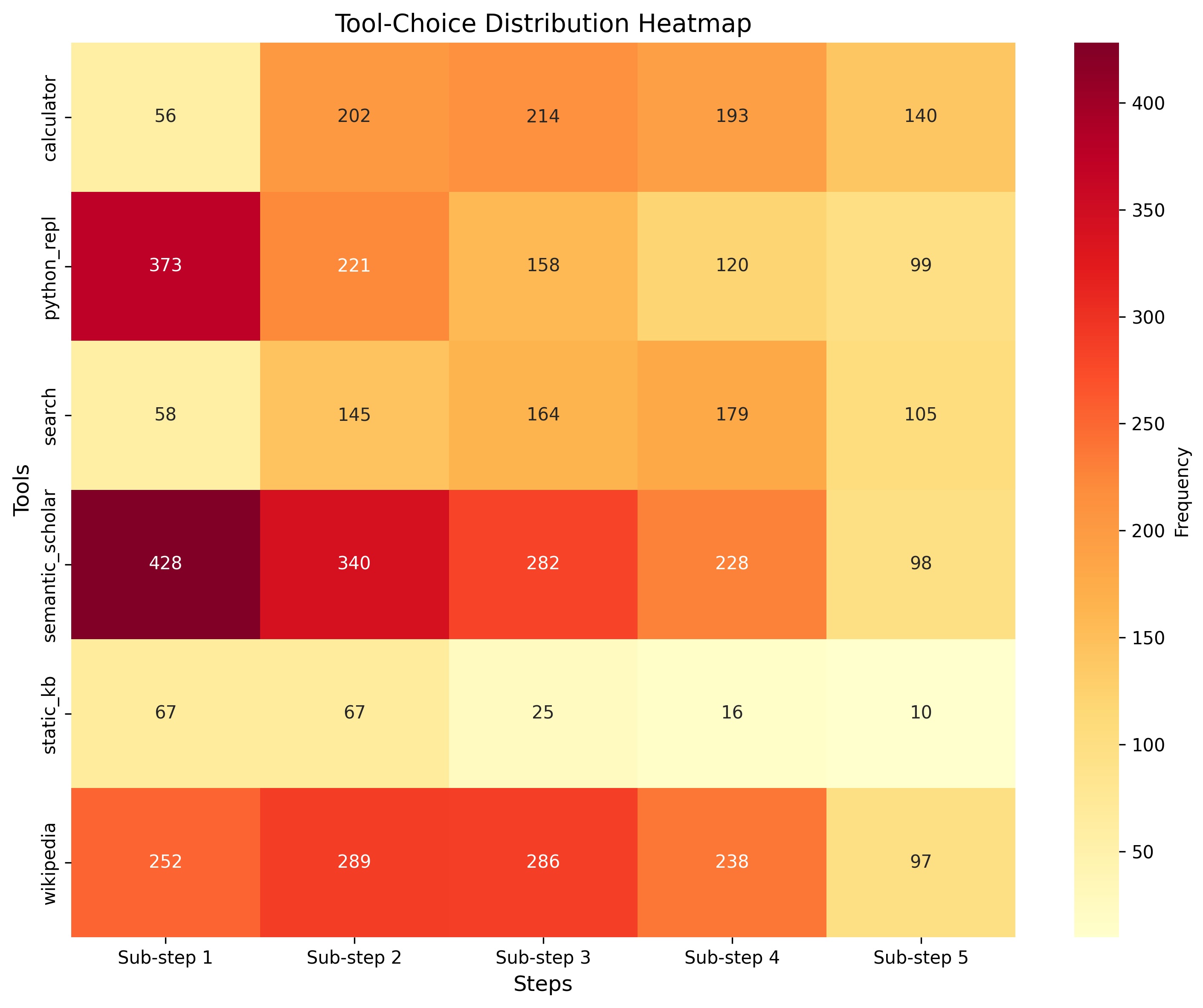} }}%
    \qquad
    \subfloat{{\includegraphics[width=0.46\linewidth]{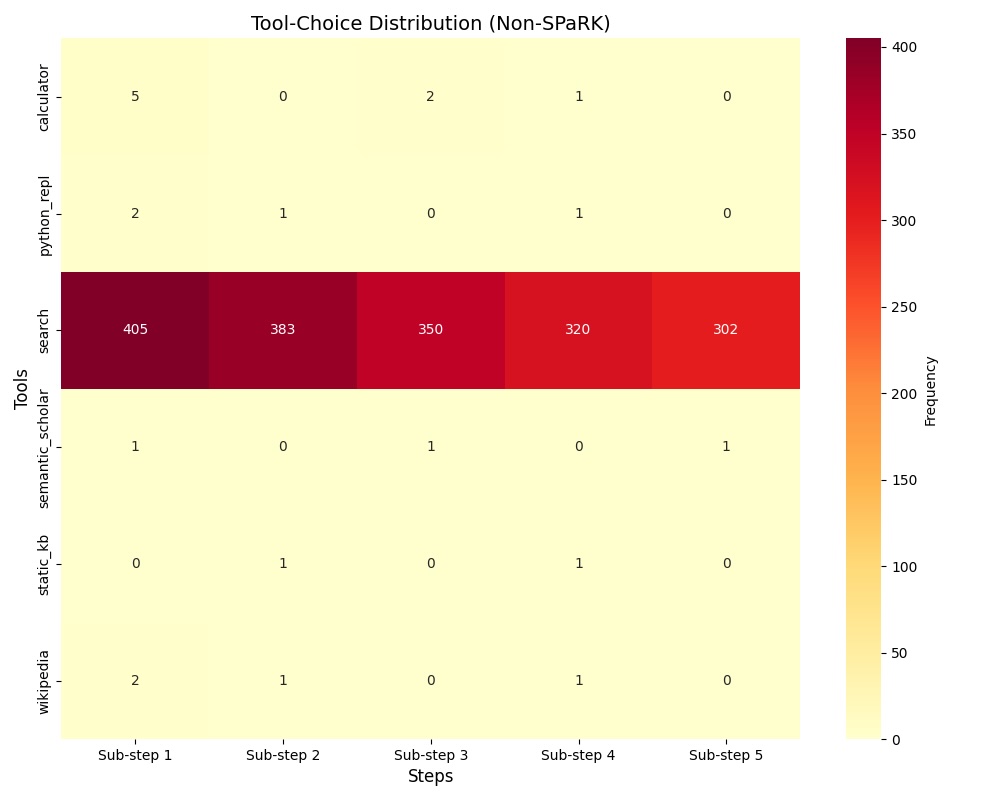} }}%
    \caption{(a) SPaRK's step-wise tool distribution displays diverse tool usage across each sub-step. \\ (b) The non-SPaRK distribution emphasizes search, even when augmented with many other tools.}%
    \label{fig:SPaRK-tool-distribution}%
\end{figure}

This behavior is reflected in the final results. The PPO model without tool diversity improves over supervised fine-tuning by a notable margin, suggesting that reinforcement learning with tools encourages better general reasoning. However, the largest gains come from the version of PPO trained with tool-diverse trajectories. The SPaRK model’s improved accuracy illustrates how structured exploration leads to substantial performance gains. These improvements are not only quantitative but behavioral. Additionally, as shown in Fig. \ref{fig:SPaRK-tool-distribution}, the SPaRK model exhibits more balanced tool usage, higher entropy in its action distribution, and increased success in categories where domain-specific knowledge is critical. This results in a policy that is both more deliberate and more effective, supporting the broader claim that tool diversity can be learned, rather than manually scripted or left to stochastic sampling.

Overall, these observations suggest that reinforcement learning with structured diverse tool trajectories offers a promising path forward for building language models that reason more like problem-solvers.

\section{Discussion}
Our experimental results reveal substantial performance improvements when employing the SPaRK framework, with the PPO-trained tool-diverse model achieving 40.8\% accuracy on MMLU-Pro compared to 26.2\% for supervised fine-tuning and 22.4\% for the baseline Llama-3.1 8B model. While these results are encouraging and suggest that reinforcement learning with explicit tool diversity rewards can significantly enhance reasoning capabilities, we acknowledge important limitations in our experimental design. Most notably, our comparison lacks a critical baseline: the base model augmented with tool access but without specialized training. This configuration would have provided a more direct assessment of whether the performance gains stem from tool availability alone or from our learned exploration policy. Such a comparison would isolate the contribution of the RL-based diversity incentives versus the mere presence of external tools, offering stronger evidence for our hypothesis that systematic tool exploration, rather than just tool access, drives improved reasoning.

Unfortunately, computational and resource constraints prevented us from conducting this additional experiment. We exhausted our allocated API credits and compute cluster credits. Additionally, time constraints in the project timeline meant we could not secure additional resources to run this baseline. Future work should prioritize this comparison, as well as explore variations in the diversity coefficient $\beta_{div}$ more thoroughly, investigate the impact of different judge models beyond GPT-4o, and extend the evaluation to other challenging datasets like HotpotQA or AIMO as originally proposed. We also believe that online PPO training, rather than our offline approach, could yield further improvements by allowing the policy to adapt dynamically to the tool selection patterns that emerge during training.

SPaRK’s design introduces a key innovation in combining exploration and exploitation during tool selection. Unlike DREAM, which splits exploration and exploitation into separate phases, SPaRK combines both objectives in a unified policy by encouraging the model to choose the least-used tool among those that still pass a quality threshold. This rarity-first strategy ensures the model avoids converging too quickly on high-frequency tools like chain-of-thought, instead learning to evaluate less familiar tools in appropriate contexts. Compared to SWiRL, which always selects the highest-scoring tool, our use of a minimum-score-above-threshold rule enables systematic exploration without discarding utility. This design leads to more diverse tool usage, which we believe is a crucial factor behind the performance improvements observed. By rewarding exploration directly within the learning objective, SPaRK helps the model uncover underutilized reasoning strategies that generalize across domains.

Beyond the immediate accuracy gains, these findings suggest a broader paradigm for retrieval-augmented and multimodal agents: tool selection itself can be cast as a learnable action space.  By baking tool‐diversity rewards into PPO, SPaRK learns a meta-retriever that adaptively decides which knowledge source or sensor to invoke—rather than relying on fixed top-$k$ search or a hard-wired modality cascade. Unlike the \citet{goldie2025syntheticdatageneration} paper that is a breakthrough in using synthetic data to generate basic samples, we are now using this recursive agentic training pipeline to do  more complex tasks such as reinforcing exploration of tooling in a model. Such a rarity-aware signal could, for example, govern when a vision–language agent should perform OCR versus symbolic math, or prompt a RAG system to pivot from a dense-passage index to a live API. This unifies hierarchical planning and language-alignment techniques under one scalable offline-RL recipe and offers a blueprint for training future tool-orchestrating LLMs whose reasoning gains stem from exploration incentives rather than parameter count alone. Future work will still have to strike a balance of how to optimize when to stop exploring the tools and exploit on the tools and chain of thought that the model deems best given the context.

\section{Conclusion}
This work introduces SPaRK, a novel reinforcement learning framework that successfully teaches language models to explore diverse tool usage patterns through explicit diversity rewards in a multi-step decision process. Our results on MMLU-Pro demonstrate that incentivizing tool exploration can almost double the performance of supervised fine-tuning approaches, achieving 40.8\% accuracy compared to 26.2\%, while maintaining high entropy in tool selection patterns. Our findings provide compelling initial evidence that algorithmic exploration through rare-tool exploitation can enhance reasoning capabilities in large language models. These results challenge the prevailing assumption that model scale and temperature-based sampling are the primary drivers of reasoning improvements, instead suggesting that principled exploration policies over heterogeneous tools represent a promising and underexplored direction for advancing language model capabilities in complex, multi-domain reasoning tasks. SPaRK highlights the potential of combining structured tool-diverse synthetic trajectories with robust RL methods like PPO to teach models not just what to think, but how to think step-by-step with tools.

\section{Team Contributions}
\begin{itemize}
  \item \textbf{Gabriel Bo:} Implemented the synthetic data generation, executed the model training, and wrote the proposal and progress report.
  \item \textbf{Koa Chang:} Implemented and trained the baseline model with and without SFT, created figures for the report, and wrote the extended abstract.
  \item \textbf{Justin Gu:} Implemented the PPO agent with actor-critic modules, designed the final poster, and wrote the final report.
\end{itemize}

\paragraph{Changes from Proposal} Several pragmatic changes were introduced during execution.  
(i) Rather than training a dedicated reward network, we employed GPT-4o directly as a lightweight reward oracle; learning a separate, distilled reward model remains promising follow-up work.  
(ii) We limited optimisation to \emph{offline} PPO (mirroring SWiRL) instead of the originally planned online updates, both to simplify tooling and to fit within API-rate constraints.  
(iii) In place of HotpotQA or AIMO we adopted the mainstream \textsc{MMLU-Pro} multiple-choice suite, whose broad subject coverage lets us measure how tool diversity impacts cross-domain reasoning.  
(iv) Finally, full-parameter SFT was deemed infeasible on a 3.1-billion-parameter Llama-3 checkpoint within available compute; we therefore confined adaptation to LoRA (and light unfreezing), leaving full SFT on a smaller architecture such as Qwen or Bert as future work.

\bibliographystyle{ACM-Reference-Format}
\bibliography{reference}

\appendix



\end{document}